\documentclass[runningheads]{llncs}
\usepackage[T1]{fontenc}
\usepackage{graphicx}
\usepackage{url}

\begin{document}
\title{How Artists Improvise and Provoke Robotics}
%
%
\author{Steve Benford\inst{1}\orcidID{0000-0001-8041-2520} \and
Rachael Garrett\inst{2}\orcidID{0000-0002-4162-9206 } \and
Eike Schneiders\inst{1}\orcidID{0000-0002-8372-1684} \and
Paul Tennent\inst{1}\orcidID{0000-0001-6391-0835} \and
Alan Chamberlain\inst{1}\orcidID{0000-0002-2122-8077} \and
Juan Avila\inst{1}\orcidID{0000-0002-5029-8541} \and
Pat Brundell\inst{1}\orcidID{0000-0002-9829-7425} \and 
Simon Castle-Green\inst{1}\orcidID{0000-0003-0681-2555}
}
\authorrunning{S Benford et al}
%
\institute{Mixed Reality Lab, The University of Nottingham, Nottingham, UK\\
\email{firstname.lastname@nottingham.ac.uk} \and
Media Technology and Interaction, KTH, Royal Institute of Technology, Stockholm, Sweden\\
\email{rachaelg@kth.se}}

\maketitle              
\begin{abstract}
We explore transdisciplinary collaborations between artists and roboticists across a portfolio of artworks. Brendan Walker's Broncomatic was a breath controlled mechanical rodeo bull ride. Blast Theory's Cat Royale deployed a robot arm to play with a family of three cats for twelve days. Different Bodies is a prototype improvised dance performance in which dancers with disabilities physically manipulate two mirrored robot arms. We reflect on these to explore how artists shape robotics research through the two key strategies of improvisation and provocation. Artists are skilled at improvising extended robot experiences that surface opportunities for technology-focused design, but which also require researchers to improvise their research processes. Artists may provoke audiences into reflecting on the societal implications of robots, but at the same time challenge the established techno-centric concepts, methods and underlying epistemology of robotics research.

\end{abstract}
\section{Introduction}

Creative Robotics \cite{hooman2023creative} brings together artists and robots in disciplines as diverse as drawing \cite{gomez2021robot}, performing music \cite{hoffman2010shimon}, opera \cite{Jochum2024opera}, theatre \cite{jochum2016using} and especially dance which has been a fascination for the community for decades ~\cite{jochum2019tonight,apostolus1990robot,maguire2024other,Grunberg;2009:DancingRobot,Tanaka:2006:DancingWithQRIO,Michalowski:2007:DancingWithKeepon,Rogel:2022:DancingWithPanda,Jochum:2019:Dance}. 
It also shapes robotics research and increasingly invites us to consider how “transdisciplinary collaborations between artists and engineers can open up new pathways for designing interactive systems” \cite{gomez2021robot}.  

We reflect on a portfolio of artworks that we co-created with artists over more than a decade---The Broncomatic (2010-2012), Cat Royale (2022-2024), and Different Bodies (2023-present)]---to explore the benefits and challenges of such collaborations, focusing on two key artistic strategies---\textit{improvisation} and \textit{provocation}. We reveal how our artistic partners employed their embodied skills and expertise to improvise aesthetic interactions with off-the-shelf robotic systems. We also reveal how artistic provocation fostered public reflection on societal concerns and challenged the ethical, conceptual, methodological, and epistemological foundations of robotics research. 

\section{Portfolio}

We follow the method of \textit{Performance-led Research in the Wild} in which researchers collaborate with artists to deliver new performances; study their rationale, processes and audiences’ responses; and from this generalise design knowledge, tools and methods~\cite{benford2013performance}. This is an example of the more general approach of Research Through Design in which knowledge emerges from design practice~\cite{zimmerman2010analysis}, including reflecting across portfolios of practice-led works~\cite{gaver2012should}. 

This paper reflects across multiple artworks in a portfolio to distil common principles that, with hindsight, can be seen to underpin them all. The projects in question arose from partnerships between the Mixed Reality Laboratory at the University of Nottingham and three different artistic partners. They were initially conceived by the artists, while the research team then helped develop and study them. Descriptions and studies of the individual projects have been published previously; what is new here are our reflections on how they invoked the two artistic strategies of improvisation and  provocation.

\subsection{The Broncomatic}
\begin{figure}[h]
\includegraphics[width=\textwidth]{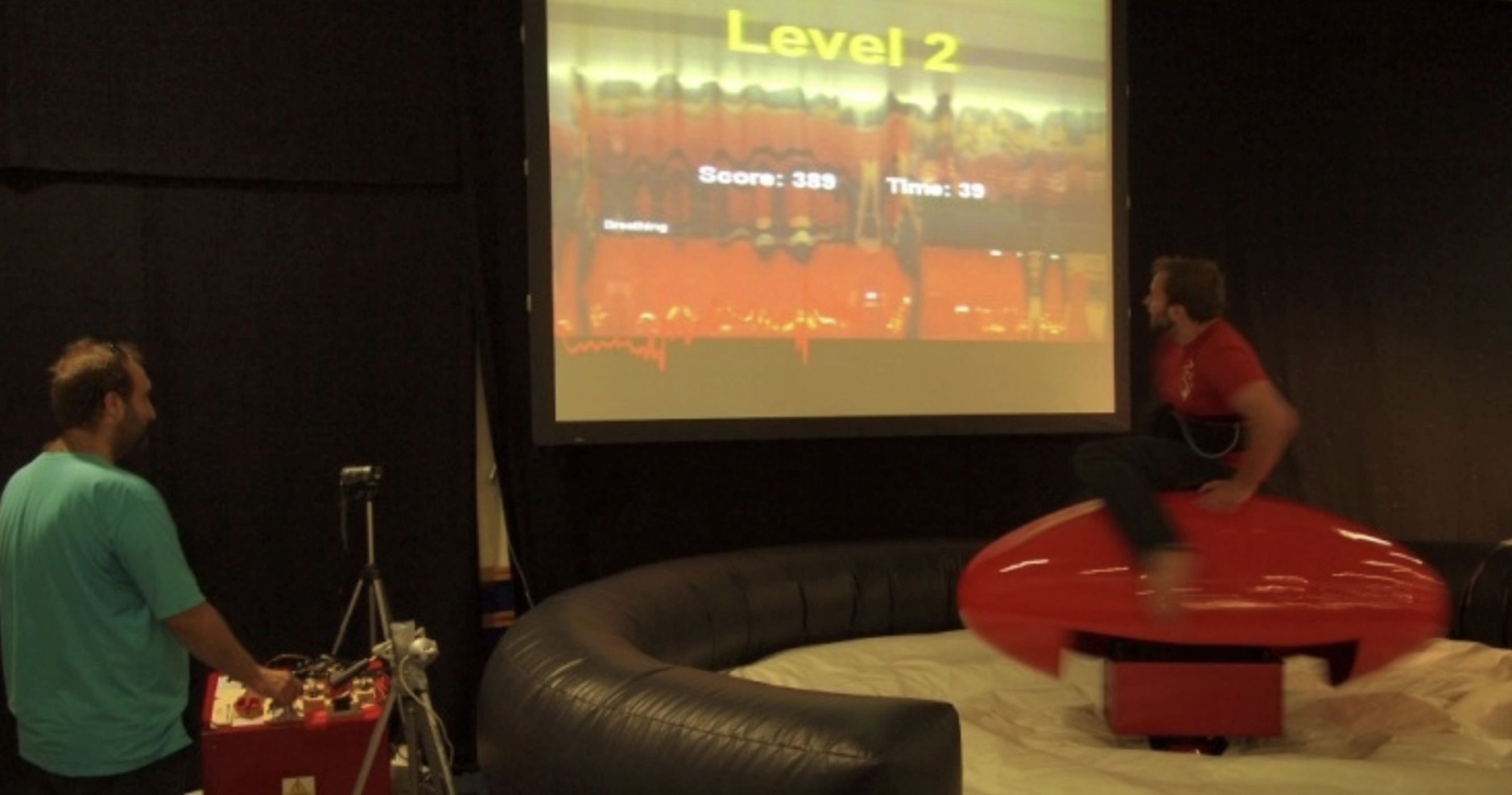}
\caption{Broncomatic rider progressing to level 2 of the thrill ride.} \label{Bronco}
\end{figure}

The Broncomatic\footnote[1]{The Broncomatic (\url{https://youtu.be/_8_IhXLtESs})} (Figure \ref{Bronco}) was an interactive thrill ride developed in partnership with the artist and ‘thrill engineer’ Brendan Walker. The team extended an off-the-shelf mechanical rodeo bull ride with a breath control interface~\cite{marshall2011breath}. The underlying ride programme was automated, gradually increasing in difficulty. However, the ride's horizontal rotation (specifically its yaw) was influenced by breathing as measured by an expandable chest strap sensor. Riders were informed that the more they breathed the more points they would score, setting up a playful dynamic in which they needed to breathe more to score points and less to stay on the ride. 

\begin{figure}[h]
\includegraphics[width=\textwidth]{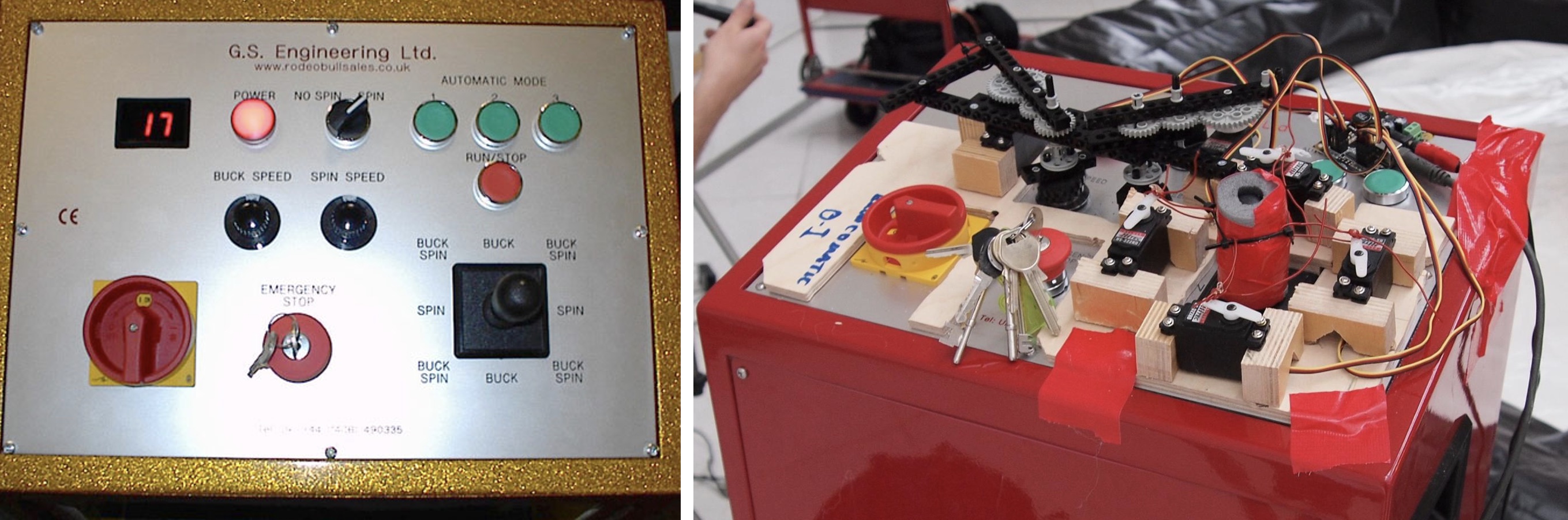}
\caption{\textit{Left:} The Broncomatic control interface. \textit{Right:} Using Lego Mindstorm technology to safely manipulate the controls.} \label{Bronco-controls}
\end{figure}

A notable feature of the Broncomatic was the robotic manipulation system shown in Figure \ref{Bronco-controls} that physically worked the ride’s controls as a human would. Using the Lego Mindstorms technology, a framework of manipulators was attached to the movement joystick, ‘spin speed’ and ‘buck speed’ controls so that the ride could be automated without having to hack into its internal control system. This improvisation was partly about ease of implementation, but was mainly driven by safety requirements; we did not need to alter the manufacturer's control system and the ride could not be moved in ways that violated its safety constraints. Note that the human standing near the control box in Figure \ref{Bronco} is solely there to monitor the ride and press the red stop button in case of emergency.

Studies of riders’ experiences of the Broncomatic, drawing on interviews and analysis of breathing data, revealed how humans contested control with the machine as part of the experience of thrill, both surrendering control to the machine and their own autonomic breathing response, and losing awareness of control during the flow of the experience~\cite{benford2021contesting}.

\subsection{Cat Royale}
\begin{figure}
\includegraphics[width=\textwidth]{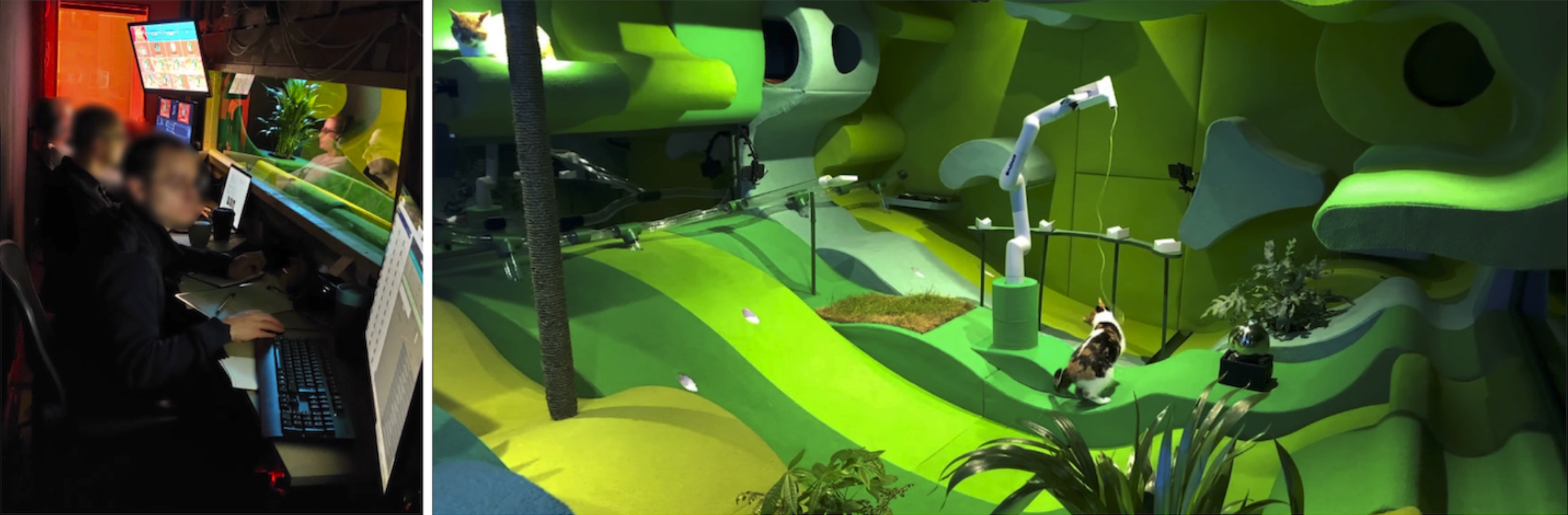}
\caption{\textit{Left:} Control room with view into the environment through one-way mirrors. \textit{Right:} Clover playing with a piece of string offered by the robot.} \label{Cat}
\end{figure}

Blast Theory's Cat Royale\footnote[2]{Cat Royale  (\url{https://youtu.be/sl6nr8B5jqQ})} engaged public audiences in reflecting on trust in robots by creating a so-called ‘cat utopia’, the purpose designed enclosure shown in Figure \ref{Cat} right that housed a family of three cats for six hours a day for twelve days, at the centre of which sat a robot arm that tried to increase their happiness by playing games with them~\cite{schneiders2024designing}. 

The robot arm, a Kinova Gen3 lite, was chosen due to a light payload and minimal range, thereby reducing risk of potential harm to the cats. It was programmed to pick up toys from nearby racks and move them in ways that would attract the attention of the cats and engage them in play. Behind the scenes, a robot operator monitored the robot, holding down a deadman’s switch for safety, a trained cat welfare officer monitored the cats’ wellbeing, a vision mixer edited footage from 8 cameras inside the environment, and the artists decided what games should be chosen (Figure \ref{Cat}, left). The artists were guided by an AI decision engine that recommended a new game every ten minutes and that gradually learned the cats’ preferences by being fed with data on the effect of each prior cat-robot encounter, especially the  impact it had on happiness which the artists measured by completing the recognised Participation in Play scale for cats after each game~\cite{ellis2022beyond}. 

A study of Cat Royale revealed how the cats played extensively, and increasingly physically, with the robot, and how the artists needed to embed the robot within a carefully designed multispecies robot world that provided opportunities for the cats to both avoid and approach the robot. The interior design of the robot world was tailored for humans, cats and computer vision, providing ample opportunities for observing and intervening from behind the scenes~\cite{schneiders2024designing}.

\subsection{Different Bodies}
\begin{figure}
\includegraphics[width=\textwidth]{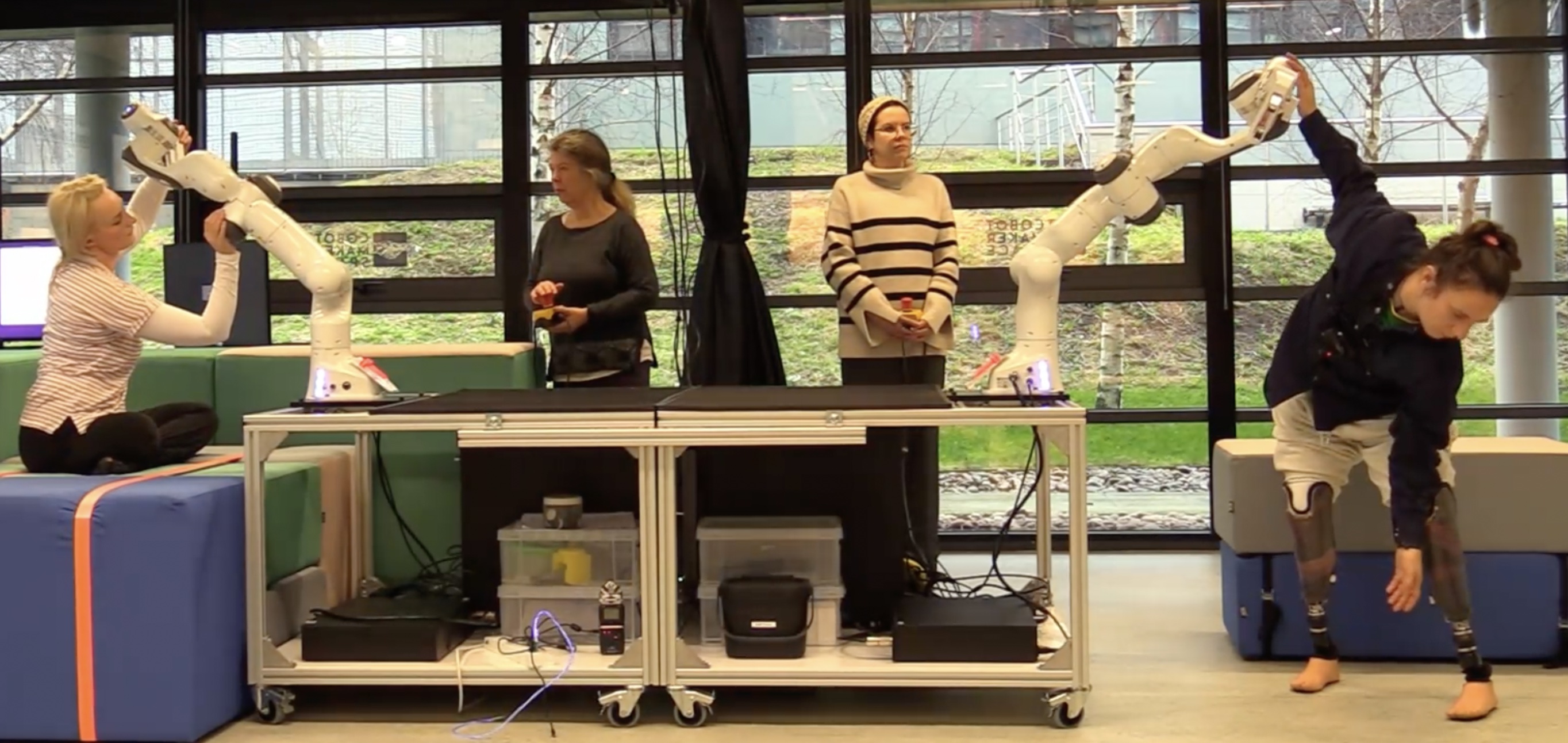}
\caption{Dancers with different disabilities dancing in harmony with a pair of Franka robot arms. To ensure safety, each robot is independently monitored and equipped with an emergency stop button for immediate cessation of operations if needed.} \label{Dance}
\end{figure}
Different Bodies\footnote[3]{Different Bodies (\url{https://youtu.be/6H28kVcrkQk})} is a collaboration between the Coventry University’s Centre for Dance Research, Candoco Dance Company, the University of Nottingham, and KTH Royal Institute of Technology to explore embodied trust in robots through a process of bringing expert moving bodies (dancers) into harmony with robots. The creative process driven by a team of professional dancers with different disabilities who may sometimes dance with assistive technologies and/or prosthetics in their practice. The motivation is to re-imagine bodily contact with robots as being creative, expressive and trustworthy rather than being a problem.

The project emerged from a previous collaboration in which dancers with disabilities danced with algorithms to generate personalised designs for aesthetic prostheses ~\cite{zhou2023beyond}. The current project focuses on dancing with robots and has so far comprised five joint co-creative workshops, where the team employed choreographic and body-based methods, including contact improvisation and soma design ~\cite{Hook2018soma}, to explore different ways of moving and dancing with robots. In early workshops, our dancers and researchers explored the aesthetic potential of various robot platforms including a Double 3 Telepresence robot, Boston Dynamics Spot, Interbotix Locobot WX250, and Franka Emika Panda robotic arm. Their improvisations guided the research process, and eventually, orientated us towards the Franka arm as being aesthetically interesting due to the expressive quality of its movement, the potential for physical engagement, and a resonance with the dancers’ perceptions of their own bodies and prostheses. This led us to establish a configuration of two Franka arms, connected together with a variety of modes, including `independent free-play’, `record and playback’, and `mirroring’, a technique in which the designated follower robot replicates the real time movements of the leader robot (Figure \ref{Dance})\footnote[4]{We thank Dr. Joseph Bolarinwa for the use of his mirroring software.}. Mirroring enabled the dancers to improvise routines in which one takes control of a robot arm, manipulating it with their hands and body, while the other improvises around the movement of its connected twin. 

Safety was a critical concern throughout, and the team gradually established a shared practice for safe and expressive bodily contact, including knowledge of how and where to touch the robots, assigning observers who could shut them down, and reorganising the physical space. Reflection on the dancers' bodily engagements with the robots yielded insights into negotiating vulnerability. On one hand, despite initial perceptions that disabled dancers might be especially vulnerable, it became clear that they were highly skilled at managing risk. On the other, despite initial perceptions of the robots being powerful and dangerous, there was a emerging sense of their  vulnerabilities, evident in freezing behaviours when pushed towards physical limits. Such insights suggested  possibilities for designing more fluid negotiation of vulnerability between human and robot.

\section{Reflections}

Our reflections turn to two aspects how our artists shaped robotics: improvisation and provocation. While both are to some extent familiar to creative robotics, we unpack subtleties of how they impact research, from the design of robot interactions to  conceptual, methodological and epistemological foundations.

\subsection{Improvisation}

Reflecting on our various projects, we highlight how artists engage in improvising experiences with robots. We emphasise the term \textit{experience} here. Our artists were looking beyond designing specific interactions with robots to instead consider extended user experiences that were aesthetic, emotional and evoked meaning making, reflecting the goals of ‘third wave’ design-oriented Human-Computer Interaction  \cite{wright2008aesthetics}. They worked with off the shelf robotic hardware (the mechanical rodeo bull and various robot arms). However, in order to deliver fully functional and robust experiences for their audiences, they needed to improvise many additional processes and implement various technologies and systems to support this.

For example, our artists improvised a control system for the Brocomatic that bypassed the need to alter its internal control system. Cat Royale relied on a robot operator who triggered and monitored the robot arm’s pre-programmed moves and could intervene to improvise new moves (e.g., when it became tangled up in the toys or when the cats played tug of war and put the arm under stress). The artists manually scored how the cats played in each game, feeding the results into the decision engine which would then recommend new games, which the artists in turn might accept or veto. Different Bodies involved two dancers improvising robot movements for one another by directly manipulating the control arm for its twin to mirror, either through the live connection between the robot arms or by being able to record and playback the robot's sequence of movements. This project also involved improvisations on the part of the roboticists to facilitate and implement new interaction possibilities---such as working with the dancers to adapt the robots to better support their movements.

From a technical perspective, these improvised systems needed to be applied at a level of scale and rigour that could deliver a fully functional and robust experience for public audiences. Public performance therefore provided a crucible for understanding the challenges that must be tackled if a robot is to successfully operate ‘in the wild’. Our artists' improvisations revealed new challenges to be solved, or requirements for future automated systems and control interfaces. This ‘experience first’ (rather than ‘technology first’) approach also reflects how the real-world deployment of robots often requires human operators and wranglers working behind the scenes to oversee and control them, i.e., that robots need to be designed as socio-technical systems with extensive human involvement required to improvise solutions to unpredictable sitations.

Improvisation also extends to research. Each of these projects produced one or more research papers. However, the themes and contributions of these papers emerged from an exploratory and open ended artistic process. Research questions were typically not identified in advance, there were no hypotheses or ‘missions’ to be tackled. For example, the idea of working with cats did not emerge until around nine months into the project that was eventually to become Cat Royale. Resulting contributions on designing multispecies robot worlds were therefore completely unanticipated at the beginning. While potentially productive, this improvisational artist-led approach to research is also challenging to funding bodies (Cat Royale was for example mainly funded as an example of public engagement for research), to ethical review boards (Cat Royale followed an eighteen month journey through three boards \cite{benford2024charting}), and to reviewers of our papers who sometimes contested the epistemological nature of our approach that led to our findings.

\subsection{Provocation}

Our experience taught us that a key role for artists in research is to provoke: to disrupt existing thinking or practice and so surface new perspectives. The value of artists in provoking public debate is already recognised within social robotics research. Jochum et al. employed live theatre to provoke audiences to consider future care scenarios between humans and robots \cite{jochum2016using},  Ocnarescu and Cossin revealed how artists may surface aesthetic and ethical challenges \cite{ocnarescu2019contribution}, and Granjon proposed that robotic art “can be a vector for techno-critique with a social impact agenda.” \cite{granjon2022robotic}. There is much public concern about the impact of robots on society, and about trust, safety, fairness and bias in relation to AI in general. With this in mind, we point towards four ways that artistic-led provocation can contribute to future robotics research.

\paragraph{Ethical:} Artists can contribute to ethical discussions, both in public and academic forums, by providing powerful first-hand experiences of robots that raise questions about future impacts. They can do so by adopting various stances towards robots (and AI more generally), from viewing them as tools for creating art, to being co-creators, to being the subject of critical inquiry~\cite{salimbeni2024decoding}. Moreover, a given artwork may combine stances as artists both embrace a technology to create an artwork while also being skeptical or critical about it. This ambiguity is explicit within Cat Royale, that has the dual goals of creating a robot to play with cats while simultaneously provoking the audience to consider whether they would trust robots to look after their loved ones, or more generally whether it would benefit humans to live in a ‘utopia’ in which robots could meet all their needs. Similarly, Different Bodies explores the potential normative assumptions made in robot design while, at the same time, employing the robots as a lens to examine a diversity of human bodies, their capabilities and vulnerabilities, and potential future relationships to technologies, materiality and autonomy. 

\paragraph{Conceptual:} Artistic provocation extends to the conceptual foundations of HRI research. Mainstream thinking about robots is that they should be dependable, safe, transparent and explainable. While these ideas are perhaps so familiar as to seem like common sense, they actually reflect a particular underlying world-view of how humans should experience robots and AI in the future. A key role for artists is to challenge such concepts by engaging robots with the messy and ambiguous world of human experience~\cite{benford2023five}. The boundaries of play and safety are re-negotiated in The Broncomatic and Cat Royale, leading us to consider the extent of our freedom to learn about and experience the world through our bodies. There is a rich history of artists employing ambiguity, deliberately creating situations that are open to multiple interpretations so as to provoke meaning-making rather than trying to give transparent and unambiguous accounts of the world~\cite{gaver2003ambiguity}. This allows for re-conceptualisations of our taken for granted concepts. Our improvisation-led research with robots led to innovative approaches to safety. While  safety remains paramount to our practice, the use of the robots for something outside their intended, engineered, purpose has required us to innovate new methods for doing the work safely: we used Lego Mindstorm to control the Broncomatic; we employed a dead man's switch for Cat Royale combined with human manipulation of the the arms; and we established a socio-technical safety framework in Different Bodies.

\paragraph{Methodological:} Provocations made by artists can also be methodological. A recent survey of AI-generated artworks showed how artists routinely subvert established AI methodologies by introducing ambiguity throughout the machine learning pipeline of dataset curation, model training, and application~\cite{sivertsen2024machine}. The same methodological provocation can be seen in our portfolio as artists subvert the established engineering and methodologies of creating robots for their own purposes. In Different Bodies, for example, our dancers probed the constraints of the Franka Arm for creative purposes---often challenging the sensibilities of others in the room. 

We reflect how the robots' various technological limitations were not always obstacles to be overcome, removed, or solved, but rather were often embraced (sometimes literally) for creative purposes. We find this a fruitful approach while working with constraints of (many) existing robotic technologies. Here, the artists prompted us to continue with the imperfect technology, adapting to it with creative and generative intention, rather than jumping to alternative design methods or solutions. 

We note how artists would often monitor and control the experience from behind the scenes, most obviously and intensively in Cat Royale. This mirrors the popular Wizard of Oz technique in which aspects of a system's operation are simulated by a human hidden behind-the-scenes ~\cite{riek2012wizard}. However, their involvement extends far beyond initial design and prototyping, with them continuing to deliver these functions in the ultimate production as they work to orchestrate the final experience. We note that such extensive orchestration work is typical of theatrical productions where large human production crews are often required to deliver a performance, making it appropriate to consider how humans can similarly become an integral part of the overall experience when designing robotic artworks.
We suggest that such an approach ultimately led to richer insights into the technology than may have been revealed through methods such as Wizard of Oz for early low-fi prototyping. We noted that, during Different Bodies, our dancers were resistant to resorting to Wizard of Oz, despite the highly constrained and limited Franka arm. 

\paragraph{Epistemological:} Finally, underlying these various ethical, conceptual and methodological provocations lurks a deeper epistemological point. We argue that the field of robotics is rooted in the largely positivist epistemology of science and engineering---in simple terms, that the world is classifiable and that there are correct solutions to given problems. In contrast, art is rooted in the subjectivist and interpretivist epistemologies of the arts and humanities, in which knowledge is subjective and it is important to recognise and even celebrate one's positionality. However, we do not see scientific and artistic research approaches as mutually exclusive endeavors. Rather, from our portfolio of artworks and their emergent knowledge contributions, we argue that these enquiries are highly entangled and serve to augment each other. We see a generative form of epistemics, wherein artists generate questions of epistemic import for robotics research. Further, these interdisciplinary collaborations facilitated rich dialogues between different experts and practitioners about how different futures with robots could be realised. In doing so, we arrived at concrete knowledge contributions for robotics. To truly engage artists with robotics research, we call for embracing their thinking alongside their methodologies, emphasising positionality and subjectivity relating to how we design interaction with robots whilst embracing artistic research practices alongside scientific ones. 


\section{Conclusion}

Artists are often highly skilled improvisers and provocateurs, and these can be powerful strategies for shaping robotics research. Improvisation can inspire design ideas for future technologies and show how robots can be embedded into wider user experiences. Provocation can engage public audiences with the moral and societal challenges arising from social robots. However, we must recognise that these strategies are a ‘two way street’ in the sense that researchers engaging with artists in transdisciplinary collaborations should be prepared to improvise around established research methodologies and expect provocation to challenge the conceptual and epistemological foundations of their own research.

\begin{credits}
\subsubsection{\ackname}
This work was supported by the Engineering and Physical Sciences Research Council through the Turing AI World Leading Researcher Fellowship in Somabotics: Creatively Embodying Artificial Intelligence [grant number APP22478], UKRI Trustworthy Autonomous Systems Hub [EP/V00784X/1], Horizon Digital Economy Hub at the University of Nottingham [EP/G065802/1], and (HoRRIzon) AI UK: Creating an International Ecosystem for Responsible AI Research and Innovation (RAKE) [EP/Y009800/1]. We thank Dr. Joseph Bolarinwa for his software to mirror the Franka robotic arms as part of the Dancing with Robots projects. We are grateful to our dancers Welly O'Brien and Kathleen Hawkins and to our artists Blast Theory, Brendan Walker and Candoco for their partnership.

\subsubsection{\discintname}
The authors have no competing interests to declare. 
\end{credits}

%
%
%

\bibliographystyle{splncs04}
\bibliography{main.bib}

\end{document}